\documentclass[conference,letterpaper]{IEEEtran}

\usepackage[utf8]{inputenc} 
\usepackage[T1]{fontenc}
\usepackage{url}
\usepackage{ifthen}
\usepackage{cite}
\usepackage{graphicx} 
\usepackage{subfigure}
\usepackage{stfloats}
\usepackage[cmex10]{amsmath} 

\usepackage{graphicx} 
\usepackage{amssymb,amsfonts} 
\usepackage[normalem]{ulem} 
\usepackage{algorithm}
\usepackage[noend]{algpseudocode}
\usepackage{enumitem}
\usepackage{xcolor}

\IEEEoverridecommandlockouts

\newtheorem{theorem}{Theorem}

\newtheorem{remark}{Remark}

\begin{document}
\title{Does Feedback Help in Bandits with Arm Erasures?} 


\author{\IEEEauthorblockN{Merve Karakas$^\dagger$, Osama Hanna$^\ddagger$, Lin F. Yang$^\dagger$, and Christina Fragouli$^\dagger$\\ 
$^\dagger$University of California, Los Angeles, $^\ddagger$ Meta, GenAI\\
Email:\{mervekarakas, ohanna, linyang, christina.fragouli\}@ucla.edu}\thanks{The work was supported in part by NSF grants \#2007714 and \#2221871, by Army Research Laboratory grant under Cooperative Agreement W911NF-17-2-0196, and by the Amazon Faculty Award.}}

\maketitle


\begin{abstract}
   We study a distributed multi-armed bandit (MAB) problem over \textit{arm erasure} channels, motivated by the increasing adoption of MAB algorithms over communication-constrained networks. In this setup,  the learner communicates the chosen arm to play to an agent over an erasure channel with probability \(\epsilon\in[0,1)\); if an erasure occurs, the agent continues pulling the last successfully received arm; the learner always observes the reward of the arm pulled. In past work, we considered the case where the agent cannot convey feedback to the learner, and thus the learner does not know whether the arm played is the requested or the last successfully received one. In this paper, we instead consider the case where the agent can send feedback to the learner on whether the arm request was received, and thus the learner exactly knows which arm was played.
    
   Surprisingly, we prove that erasure feedback does not improve the worst-case regret upper bound order over the previously studied \emph{no-feedback} setting. In particular, we prove a regret lower bound of \(\Omega(\sqrt{KT} + K/(1-\epsilon))\), where \(K\) is the number of arms and \(T\) the time horizon, that matches no-feedback upper bounds up to logarithmic factors. We note however that the availability of feedback does enable to design simpler algorithms that may achieve better constants (albeit not better order) regret bounds; we design one such algorithm, and numerically evaluate its performance.  
\end{abstract}

\section{Introduction}
\label{sec:intro}

The multi-armed bandit (MAB) framework has emerged as a fundamental model for sequential decision-making under uncertainty, finding applications in areas such as recommendation systems, clinical trials, distributed robotics, and online advertising \cite{9185782}. In the classic MAB problem, a learner selects (or \emph{pulls}) one of $K$ arms in each round, observes the associated reward, and aims to maximize cumulative rewards over a time horizon  of $T$ rounds. The problem inherently involves balancing the exploration of unknown arms with the exploitation of arms believed to yield higher rewards, a trade-off encapsulated by the cumulative regret metric \cite{lattimore2020bandit}.

While classical MABs are well studied, extending the framework to \emph{distributed} environments significantly increases complexity. Communication constraints, channel imperfections, and incomplete feedback all introduce new challenges.
In this work,  we consider a scenario where
 environmental interference or hardware constraints can cause \emph{arm erasures}. This captures real-world settings such as:
\begin{itemize}
    \item \textbf{Drone traffic management.} A central learner coordinates maneuvers for drones via lossy wireless channels. The learner observes drone performance through sensors or video streams~\cite{hosny2023budgeted, 9367568}.
    \item \textbf{Medical microrobotics.} Nanobots operating inside the human body rely on limited and often error-prone communication links to receive navigation commands~\cite{abbasi2024autonomous, yang2024machine}.
\end{itemize}

\begin{figure}[t b!]
  \centering
  \includegraphics[width=0.8\linewidth]{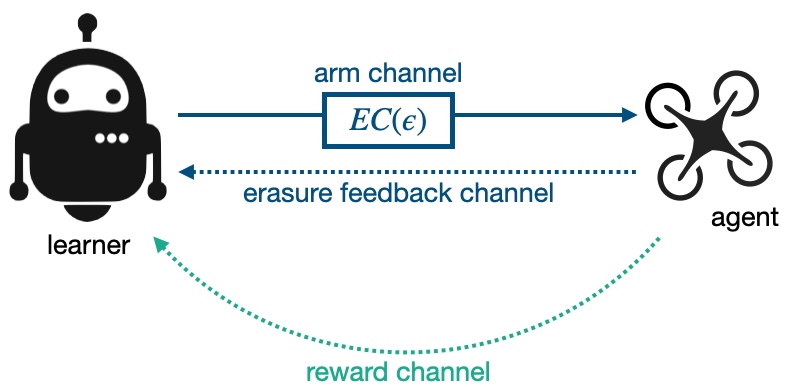}\
  \caption{The figure illustrates the communication structure between the learner and the agent in a multi-armed bandit system with erasure feedback. The learner selects an arm to be played and sends it to the agent via the arm channel, which is modeled as an erasure channel $\text{EC}(\epsilon)$ with erasure probability $\epsilon \in [0, 1)$. The learner observes the reward and erasure feedback indirectly (through reliable sensors).}
  \vspace{-.2in}
  \label{fig:learner-agent}
\end{figure}

In particular, we consider the scenario depicted in Fig.\ref{fig:learner-agent} where a learner at each round communicates the chosen arm to play to an agent over an erasure channel with probability $\epsilon \in [0, 1)$; if an erasure occurs, the agent continues
pulling its last successfully received arm; the learner always observes the reward of the arm pulled.
 In past work, we considered the case where the agent cannot convey feedback to
the learner, and thus the learner does not know whether the arm played is the requested or the last successfully received one \cite{singleagentprior}.
As a result, the learner observes a reward but does not exactly know which arm this reward is associated with; the learner has only probabilistic knowledge on which arm was played. 

    In this paper, we instead consider the case where the agent can send feedback on whether the arm request was received, and thus the learner knows exactly which arm was played at each time slot. As a result, the learner can associate the observed reward with the correct arm played.  We ask: \\
    \emph{How much does it help if the learner \textbf{does} know when an arm is erased?}

In particular, we address the following research questions:
\begin{itemize}
    \item How does knowledge of erasure events (feedback) influence \emph{fundamental}  regret bounds?
    \item Can we design algorithms that leverage feedback to reduce algorithmic complexity?
    \item How do algorithms that use feedback compare empirically to existing no-feedback approaches?
\end{itemize}

Surprisingly, our work reveals that  \emph{erasure feedback} does \emph{not} improve the worst-case regret bound. Nevertheless, feedback may offer  practical advantages: it enables the design of simpler algorithms that can promptly adapt to erasure events, without resorting to the more involved offline correction or repetition strategies required in the absence of feedback. Moreover, it removes the need to know or estimate the erasure probability before running the learning algorithm, and it enables creating algorithms that can adapt to varying erasure probabilities. 

\subsection{Key Contributions}
Our main contributions are as follows:
\begin{itemize}
    \item We establish a lower bound on the regret of single-agent MAB systems with erasure feedback, demonstrating that the worst-case performance of the learner remains unchanged compared to the no-feedback scenario.
    \item We propose a novel feedback-based algorithm that leverages erased rounds to implement more efficient and practical scheduling. Our algorithm achieves near-optimal regret guarantees without requiring the knowledge of erasure probability.
    \item We validate our theoretical findings through simulations, demonstrating that while feedback does not shift fundamental performance limits, it enables to achieve performance gains with simple algorithm designs.
\end{itemize}

\subsection{Related Work}
Multi-armed bandit (MAB) problems have long been studied as a core model of sequential decision-making under uncertainty (e.g., \cite{Thompson1933ONTL, bubeck2012regret, Auer95Adv, Lai1987AdaptiveTA, hanna2024efficient} and references therein), with applications in areas such as recommendation systems, online advertising, and distributed robotics~\cite{lattimore2020bandit}. In the canonical stochastic setting, algorithms like Upper Confidence Bound (UCB)~\cite{Auer2002, Lai1987AdaptiveTA} and Thompson Sampling~\cite{Thompson1933ONTL} achieve near-optimal gap-dependent and worst-case regret bounds. Although such approaches are powerful in settings where arms are reliably pulled and reward signals are fully observed, they often fail or become suboptimal when extended to more complex scenarios involving adversarial corruption, delayed feedback, or unreliable arm transmission.

In the realm of imperfect reward observations, numerous works have studied different kinds of corruptions~\cite{Lykouris18, hanna2023contexts, pmlr-v99-gupta19a, hanna2022solving, Kapoor19ProbCorReward, hanna2022learning, hanna2022compression,  hanna2024differentially}, e.g., \emph{adversarial corruption} where the rewards themselves are partially manipulated or perturbed. For instance, Lykouris et al.~\cite{Lykouris18} introduced stochastic MABs with an adversarial corruption budget $C$ and proposed an algorithm achieving $\tilde{O}(C \sqrt{KT})$ in the worst case. Subsequent improvements by Gupta et al.~\cite{pmlr-v99-gupta19a} and Amir et al.~\cite{amir2020prediction} provide near-matching upper and lower bounds, demonstrating that even bounded corruption can significantly degrade MAB performance. Meanwhile, delayed and censored feedback has been shown to increase regret by an additive or multiplicative factor, depending on the delay model~\cite{pmlr-v99-gupta19a, Kapoor19ProbCorReward}. These works, however, primarily focus on uncertainties in the \emph{reward signal}, assuming that arms themselves are deployed as intended in each round.

In our recent work we explored \emph{arm erasures}, where the learner’s chosen arm may be ``lost'' during transmission. In this model, the learner does not observe any feedback (beyond the noisy reward) about whether an arm was delivered successfully.  In~\cite{singleagentprior}, we consider a single-agent bandit under an erasure probability $\epsilon$, proposing a repetition-based approach that adds at most $\tilde{O}\!\bigl(\tfrac{K}{{1 - \epsilon}}\bigr)$ to the regret bound of classical MAB algorithms. Extending to multiple agents, in~\cite{hanna2024multi}, we introduced channel scheduling and repetition protocols that allow a central learner to communicate with agents over heterogeneous erasure channels, achieving sub-linear regret and highlighting that naive algorithms may suffer $\Omega(T)$ (linear) regret if erasures are neglected. Furthermore, best-arm-identification (BAI) under arm erasures has been studied in~\cite{bai-arm-erasures}, where the primary goal is to minimize the stopping time subject to a fixed-confidence constraint. Despite varying objectives---cumulative regret vs.\ pure exploration---these works collectively demonstrate that arm erasures create substantial uncertainty about the executed arm, {and carefully designed algorithms are required to achieve sub-linear regret or bounded stopping times}. 
In contrast to our prior work which assumes \emph{no} erasure feedback, this paper investigates \emph{how the availability of erasure feedback influences MAB performance}. 

\smallskip
\noindent
\textbf{Paper Organization.} The remainder of the paper is organized as follows. Sec.~\ref{sec:model} introduces our system model and notation. Sec.~\ref{sec:theory} presents our theoretical results and the proposed feedback-based algorithm. Sec.~\ref{sec:experiments} describes our simulation setup and discusses the empirical performance. Sec.~\ref{sec:conclusion} concludes the paper and outlines directions for future research.

\section{Problem Setup}
\label{sec:model}
\noindent \textbf{System Model.}
We consider a multi-armed bandit (MAB) problem with \(K \ge 2\) arms, indexed by \(\{1,2,\dots,K\}\). A single agent is connected to a central learner through an \emph{erasure channel}. The learner operates over a horizon of \(T\) discrete rounds, indexed by \(t \in \{1,2,\dots,T\}\). At the beginning of round \(t\), the learner selects (and transmits) an arm \(a_t\). Because the channel is imperfect, the transmitted arm is received with probability \(1-\epsilon\), and erased with probability \(\epsilon \in [0,1)\). These erasure events are independent across rounds.

\noindent
\textbf{Erasure Feedback.}
Unlike prior works that assume \emph{no} erasure feedback at the learner, we consider a setting where the learner is \emph{immediately informed} whether the transmission is erased or not. Formally, let \(e_t \in \{0,1\}\) denote an i.i.d.\ Bernoulli random variable indicating an erasure at round \(t\); \(e_t = 1\) if the arm is erased, and \(0\) otherwise. In our setting, the learner \emph{observes} \(e_t\) at the end of each round, thus knowing whether its chosen arm was successfully delivered.

\noindent
\textbf{Agent Operation.}
If the transmitted arm at time \(t\) is not erased (\(e_t = 0\)), the agent updates its pulled arm to \(a_t\). Otherwise (\(e_t = 1\)), the agent continues pulling the arm from the previous round. Let \(\tilde{a}_t\) denote the \emph{actual} arm pulled by the agent at round~\(t\), i.e.,
\[
\tilde{a}_t \;=\;
\begin{cases}
a_t, & \text{if } e_t = 0,\\
\tilde{a}_{t-1}, & \text{if } e_t = 1.
\end{cases}
\]
If the first arm (at \(t=1\)) is erased, we assume \(\tilde{a}_1\) is chosen uniformly at random from the \(K\) arms. Crucially, we assume the \emph{learner knows this strategy}, i.e., it is aware that the agent always pulls the last successfully received arm.

\noindent
\textbf{Reward Model.}
Each arm \(i \in \{1,\dots,K\}\) is associated with an unknown reward distribution, with mean \(\mu_i\). {The collection of these \(K\) reward distributions defines a \textit{bandit instance}. We denote the gap between the mean reward of arm \(i\) and the mean reward of the optimal arm as \(\Delta_i\). To simplify the analysis, we assume that all rewards are supported on the interval \([0,1]\), ensuring bounded variance, though results extend to sub-Gaussian reward distributions.} We let \(r_t\) denote the reward obtained when the agent pulls the actual arm \(\tilde{a}_t\) at round \(t\).  Thus, at each round \(t\), the learner gathers:     
(i) reward feedback \(r_t\) and (ii) the erasure indicator \(e_t\)  
through a reliable feedback mechanism or error-free communication channel.

\noindent
\textbf{Objective: Regret Minimization.}
We define the (expected) regret as the gap between the best achievable cumulative reward and the actual cumulative reward obtained by the pulled arms. Specifically,
$$
R_T \;=\; T \max_{i \in \{1,\ldots,K\}} \mu_i 
\;-\; \mathbb{E}\!\Bigl(\,\sum_{t=1}^T \mu_{\tilde{a}_t}\Bigr),
$$
where the expectation is taken over both the random choice of arms by the learner, and the randomness of erasures and rewards. The goal of the learner is to design a policy (i.e., a sequence of arm-selection rules) that leverages erasure feedback to minimize \(R_T\) over the horizon \(T\).

\noindent
{
\textbf{Prior Setup and Results.} In our earlier work, we introduced the concept of multi-armed bandits operating over arm erasure channels with no erasure feedback \cite{singleagentprior}. We demonstrated that, under this setting, the learner must contend with the uncertainty of whether the observed rewards correspond to the requested or fallback actions. Our key results included:
\begin{itemize}[leftmargin=*, itemsep=0pt]
    \item A generic robustness layer for any MAB algorithm $\pi$,  achieving regret $O(\alpha R^{\pi}_{\lceil T / \alpha \rceil}(\{\Delta_i\}_{i=1}^K))$, where $R^{\pi}_T$ bounds the regret of $\pi$ for gaps $\{\Delta_i\}_{i=1}^K$,
    \item A modified Successive Arm Elimination (SAE) \cite{auer2010ucb} algorithm with regret bounded by $O\left(\frac{K\log T}{1-\epsilon}+\sum_{i:\Delta_i>0}\frac{\log T}{\Delta_i}\right)$ with high probability, demonstrating near-optimal performance, and
    \item A lower bound of $\Omega\left(\sqrt{KT} + \frac{K}{1-\epsilon}\right)$, establishing the fundamental limits of regret in the absence of erasure feedback.
\end{itemize}}

\noindent
\textbf{Discussion and Connections.}
When \(\epsilon = 0\), this setting reduces to a classical MAB problem without communication errors. Conversely, if the learner \emph{did not} receive the erasure indicator \((e_t)\), we recover the no-feedback erasure scenario studied in \cite{singleagentprior} and related works. Our framework thus unifies these cases while investigating how \emph{erasure feedback} influences algorithm design and achievable regret. We note that \emph{best-arm identification} (BAI) can also be considered in a related formulation by focusing on sample complexity rather than cumulative reward; however, in this work, we concentrate on the standard regret objective.

\section{Lower Bound Analysis and Proposed Algorithms}
\label{sec:theory}

This section presents our theoretical analysis of regret minimization in multi-armed bandit systems with erasure feedback. We first establish an instance-independent lower bound on the expected regret for the single-agent setting. Notably, this lower bound aligns with the existing lower bound for scenarios without feedback \cite{singleagentprior, hanna2024multi}, indicating that the introduction of erasure feedback does not enhance the fundamental performance guarantees. Our results demonstrate that while erasure feedback provides additional information to the learner, it does not lead to order-wise improvements in regret bounds. 

\subsection{Lower Bound on Single Agent with Erasure Feedback}

We prove an instance-independent lower bound for the setting with erasure feedback, where the learner is aware of the erasures and which arm the agent pulls in each round. 
\begin{remark}[Low-erasure regime]
As recalled in Sec.~\ref{sec:model}, the generic robustness scheme of
\cite{singleagentprior} guarantees
\(\mathbb{E}[R_T]=\tilde O(\sqrt{KT})\) whenever
\(\epsilon<0.5\) for any optimal MAB algorithm, e.g., UCB~\cite{Auer2002,Lai1987AdaptiveTA} or
SAE~\cite{auer2010ucb}.
This matches, up to logarithmic factors, the clean-channel lower bound \(\Omega(\sqrt{KT})\)~\cite{lai1985asymptotically,auer1995gambling}.
Hence no additional lower-bound work is needed for
\(\epsilon<0.5\); the remainder of this section focuses on the high-erasure regime \(\epsilon\ge0.5\).
\end{remark}

\begin{theorem} \label{thm:single-agent-lb-revised}
    Consider a multi-armed bandit setting with an agent connected through an erasure channel with erasure probability $\epsilon \geq 0.5$ and erasure feedback. For $K>1$, time horizon $T \geq \frac{K}{4(1-\epsilon)}$, and any policy $\pi$, there exists a $K$-armed bandit instance $\nu$ such that
    \[
        \mathbb{E}[R_T(\pi, \nu)] = \Omega\left(\sqrt{KT} + \frac{K}{1-\epsilon}\right),
    \]
    where $\mathbb{E}[R_T(\pi, \nu)]$ is the expected regret of the policy $\pi$ over the instance $\nu$.
\end{theorem}

We provide a proof sketch below, see Appx.~\ref{apx:single-agent-lb-revised} for details.

\noindent\textit{Proof Sketch.} 
\begin{enumerate}
    \item \textit{Construct $K$ distinct bandit instances.} For each arm $i\in\{1,\dots,K\}$, define an instance $\nu_i$ where arm $i$ has reward $1$ deterministically, and the other $K-1$ arms have reward $0$. If the agent ever actually plays arm $i$, the learner sees reward $1$; otherwise $0$.
    \item \textit{Define events forcing large regret.} We consider the first $\frac{K}{4(1-\epsilon)}$ rounds and look at how many times the “true best arm” is pulled. We show that with constant probability $>1/2$, the number of nonerased transmissions is so small that for at least one bandit instance, the best arm remains essentially untested or overshadowed by random arms. 
    \item \textit{Erasure feedback does not circumvent the need to keep exploring.} Even though the learner knows exactly which rounds are erased, in the worst case, the channel realization can make it \emph{impossible} for the best arm to have been pulled enough times to differentiate it. This yields a linear portion of regret $\frac{K}{4(1-\epsilon)}$. 
    \item \textit{Combined with standard $\sqrt{KT}$ lower bound.} The well-known $\Omega(\sqrt{KT})$ bound from classical MAB analysis remains. Taking $\max$ of these two yields the final $\Omega\bigl(\sqrt{KT} + \frac{K}{1-\epsilon}\bigr)$ expression.
\end{enumerate}
Thus, the direct knowledge of erasures each round does not reduce the overall scaling of regret in the worst-case scenario.

Lower bound in Theorem~\ref{thm:single-agent-lb-revised}, which assumes knowledge of erasures, matches the lower bound exactly and upper bound up to logarithmic factors presented in \cite{singleagentprior}. Consequently, it suggests that erasure feedback cannot enhance the order optimality of the single-agent systems. 

\subsection{Proposed Algorithm with Erasure Feedback}
\label{sec:proposed-algo}

Although our lower bound shows that erasure feedback does \emph{not} improve the order of the worst-case regret, it can be exploited to simplify the algorithmic design. In the no-feedback scenario, one must rely on fixed (blind) repetition—sending each arm \(\alpha\) times—to ensure with high probability that at least one transmission succeeds. Here, with erasure feedback available, the learner can \textit{stop transmitting as soon as a success is observed}, without needing to know the erasure probability \(\epsilon\) in advance. {This feature is attractive in practical settings where \(\epsilon\) may be unknown or costly to estimate.}

\smallskip
\noindent
\textbf{Algorithm Idea: \texttt{Stop-on-Success} (SoS).}  
Suppose the learner is running a base policy, a standard MAB policy (e.g., Successive Elimination \cite{auer2010ucb} or UCB \cite{Auer2002, Lai1987AdaptiveTA}), that dictates which arm to “pull” at conceptual timesteps \(1,2,\dots\). Each conceptual pull might require multiple actual transmissions {to be successfully received by the agent}. In the Stop-on-Success (SoS) approach, at each step:
\begin{itemize}
    \item The learner picks an arm \(a\) to play, and then transmits it repeatedly \emph{until} it observes a successful transmission (i.e., until it sees \(e_t = 0\)).
    \item Once success is observed, the learner knows that the agent is playing arm \(a\) and stops sending further repetitions.
    \item {After the agent plays \(a\) for that round and the reward \(r_t\) is observed, that successful pull is recorded (and the agent sets the last successfully arm to \(a\)).} 
    \item {The learner updates the base policy using \((a, r_t)\), and moves on to the next conceptual pull as per the policy.}
\end{itemize}
This procedure is reminiscent of the “Repeat-the-Instruction” approach in \cite{singleagentprior}, except that here the learner can cease transmission upon the first success. In the no-feedback case, one must set a predetermined repetition length (e.g., \(\alpha = \Omega(\log{T}/\log{(1-\epsilon)})\) times) to guarantee a success with high probability. In contrast, our SoS method removes the need to design a fixed-length repetition schedule, thereby reducing overhead in typical channel realizations.

\subsection{Specific Application: SoS with Batched Algorithms}
{We now specialize the SoS approach to batched elimination policies, in particular to the Successive Arm Elimination (SAE) algorithm~\cite{auer2010ucb}. SAE partitions the time horizon into batches whose lengths grow exponentially, with each surviving arm being pulled \(m_i = 4^i\) times during batch \(i\). At the end of each batch, arms are eliminated based on confidence intervals that shrink as more samples are gathered.}

{Under the SoS framework applied to SAE, the learner ensures that within each batch an active arm is observed for \(m_i\) consecutive pulls. This modification leverages the agent’s operation so that after the first successful transmission, consecutive pulls generate rewards without incurring extra repetition overhead.}

\begin{algorithm}
  \caption{Stop-on-Success (SoS) Algorithm for SAE}
\label{alg:SoS}  
\begin{itemize}
    \item \textbf{Initialize:} Candidate arm set \(A=[K]\); batch index \(i=1\).
    \item \textbf{For batch \(i\) (until \(T\) is exhausted):}
    \begin{itemize}
        \item The base policy \(\mathcal{\pi}\) prescribes a set of active arms \( A = \{a_1,\dots,a_J\}\) to be pulled \(m_i= 4^i\) times each.
        \item For each arm \(a\) in the set:
        \begin{itemize}
            \item \textbf{Transmit Phase:} Transmit \(a\) repeatedly until a success is observed (i.e., until \(e_t=0\)).
            \item \textbf{Collection Phase:} After success, record the rewards for \(m_i\) rounds, \(r_t, \dots, r_{t+m_i}\), for \(a\).
        \end{itemize}
        \item Compute the empirical mean \(\widehat{\mu}_a^{(i)}\) for each \(a\) from its \(m_i\) recorded rewards.
        \item Update the active set:
        \[
          A \gets \Bigl\{a\in A \,\Big|\, \max_{b\in A}\widehat{\mu}_b^{(i)}-\widehat{\mu}_a^{(i)}\leq 4\sqrt{\frac{\log(KT)}{(2 \cdot m_i)}}\Bigr\}.
        \]
        \item Increment batch index: \(i\gets i+1\).
    \end{itemize}
\end{itemize}
\end{algorithm}

\subsection{Performance Upper Bound and Discussion}

\begin{theorem}[Upper Bound on SoS for SAE]
\label{thm:sos-sae}
    Consider time horizon  $T$. For any $K$-armed bandit problem, Algorithm~\ref{alg:SoS} achieves a regret bound of 
    $$R_T = \tilde{O}\left( \sqrt{KT} + \frac{K}{1-\epsilon} \right),$$
    where $\epsilon$ is the erasure probability of the channel and $\tilde{O}$ hides the poly-logarithmic factors.
\end{theorem}

We next provide a proof sketch; see Appx.~\ref{apx:single-agent-sae-revised} for details.

{
\noindent\textit{Proof Sketch.} 
By bounding the number of rounds that takes an arm to be successfully received with high probability, and applying an analysis similar in spirit to that in our previous work \cite{singleagentprior}, one can show that the SoS approach incurs worst-case regret of $\tilde{O}\!\Bigl(\sqrt{KT} + \frac{K}{1-\epsilon}\Bigr)$ for SAE.}

In worst-case realizations, we may need to repeat the arms as many times as the no-feedback algorithms. However, if a success happens quickly, we do fewer transmissions, which leads to faster learning. 

\smallskip
In summary, the Stop-on-Success approach offers a simpler and more robust algorithmic design that can be deployed on top of any standard MAB policy and can provably provide near-optimal regret guarantees when applied on top of batched algorithms such as SAE. 

\section{Experiments}
\label{sec:experiments}

\begin{figure*}[t b!]
  \centering
  \includegraphics[width=0.9\linewidth]{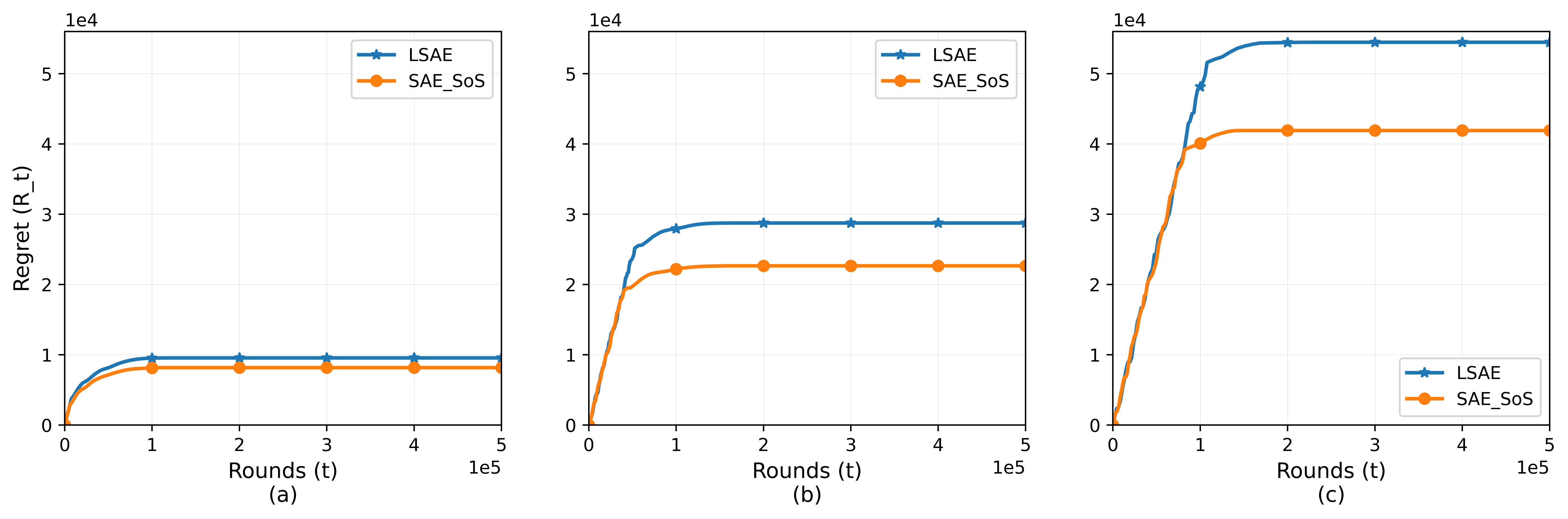}\
  \caption{Comparison of Regret over the Horizon For Different Erasure Probabilities. From Left To Right, The Plots Show Cumulative Regret As A Function Of Rounds t For erasure probabilities of (a) 0.5, (b) 0.9, and (c) 0.95, Respectively.}
   \vspace{+.05in}
  \label{fig:k10_t5e4_eps_05_09_095}
\end{figure*}

In this section, we empirically evaluate the regret performance of our proposed algorithm Alg.~\ref{alg:SoS}, SoS applied on Successive Arm Elimination (SAE\_SoS), and compare against the following method:
\begin{itemize}
    \item LSAE: This is the Lingering Successive Arm Elimination (SAE) algorithm we proposed in our previous work \cite{singleagentprior}. It operates under the assumption that there is no erasure feedback, and includes a repetition period of $\alpha = \Omega(\log{T} / (1 - \epsilon))$ before starting to collect rewards.
\end{itemize}

Our experiment, shown in Figure~\ref{fig:k10_t5e4_eps_05_09_095}, uses $K=50$ arms, with Gaussian reward distributions that have variance $1$ and means generated uniformly randomly in the range $[0,1]$.  The time horizon is $T= 10^6$ (Figure~\ref{fig:k10_t5e4_eps_05_09_095} shows until $T'= 5 \times 10^5$ to get a closer look) and the regret in each plot is averaged over $100$ experiments with arms shuffled. The channels have erasure probabilities  $0.5$, $0.9$, and $0.95$, respectively.

We observe that while the asymptotic behavior of both algorithms remains the same, SAE\_SoS achieves a constant-factor improvement in regret over LSAE. This improvement increases with the erasure probability $\epsilon$, as SAE\_SoS benefits from stopping repetitions at the first success rather than completing a fixed repetition period $\alpha = \Theta(\log{T} / \log(1/\epsilon))$.

The number of erasures until the first successful transmission follows a geometric distribution with mean $1 / (1-\epsilon)$, while LSAE performs a fixed $\alpha$ repetitions. As $\epsilon \to 1$, $\alpha  \sim \Theta(\log{T} / (1-\epsilon))$, and the difference between the mean of the distribution and $\alpha$ scales with $1 / (1-\epsilon)$, leading to significant savings in regret as $\epsilon$ increases. By leveraging erasure feedback, SAE\_SoS avoids unnecessary repetitions, making it particularly effective under high erasure probabilities.

\section{Discussion and Future Work}
\label{sec:conclusion}

Our work bridges the gap between theoretical limits and practical algorithm design in single-agent MAB settings, offering insights into the interplay between feedback and algorithmic performance. One might expect that direct knowledge of whether an arm reached the agent would yield significant gains in regret. Surprisingly, we prove that the fundamental worst-case performance remains unchanged, showing that erasure feedback \emph{does not} improve the order-optimal regret, although 
we demonstrate in numerical evaluation performance benefits. {We note that we can achieve these benefits using a minimal (one-bit) feedback  per round, which can suit well practical settings and requirements. For instance:

\begin{itemize}
    \item \textbf{Resource-limited systems.} Devices such as low-power sensors or microrobots may not support high-rate uplink but can still transmit a simple binary acknowledgment. 
    \item \textbf{Reprogrammable agents.} An agent can be equipped with low-complexity acknowledgments prior to deployment, reducing energy and design overhead compared to full data uplink.  
    \item \textbf{Reduced transmission overhead.} Once an arm is acknowledged, the SoS approach ceases further transmissions for that pull, reducing the overhead, crucial in shared channels with strict energy or frequency constraints.   
    \item \textbf{Adaptability.} While we model the channel with a fixed erasure probability, real-world channels often vary in quality over time with an estimate of the mean erasure rate. The SoS approach naturally adapts to such variations, dynamically adjusting to channel conditions without relying on a fixed erasure probability, ensuring robust performance even in fluctuating environments.
\end{itemize}}

Future work includes  exploring setups with multiple agents and adversarial erasures. A natural next step is to consider a system with $M$ homogeneous agents, all having the same erasure probability. We expect the design principles behind our single-agent feedback strategies to carry over to the multi-agent setting, although coordinating transmissions among agents can introduce new challenges in the analysis.  
Another intriguing direction is to study an adversarial erasure model, where an adversary may choose which transmissions fail. Understanding how robust our approach is under adversarial corruptions of arm transmissions is an open question.

\newpage
\IEEEtriggeratref{15}
\bibliographystyle{IEEEtran}
\bibliography{references}




\onecolumn 
\appendices 

\section{Proof of Theorem~\ref{thm:single-agent-lb-revised}}
\label{apx:single-agent-lb-revised} 

\noindent\textit{Proof.}  
A lower bound of $\Omega(\sqrt{KT})$ on the worst-case regret is already proved in literature, see \cite{lattimore2020bandit} and references therein; hence, we prove a lower bound of $\Omega(K/(1-\epsilon))$. We consider $K$ bandit instances, each with $K$ arms, where in instance $\nu_i$ the means of the reward distributions are
\begin{equation}
    \mu_j^{(i)} = \mathbf{1}\{j=i\}, \quad j=1,\ldots,K
\end{equation}
with $\mathbf{1}$ the indicator function. We assume a noiseless setting where picking arm $j$ in instance $\nu_i$ results in reward $\mu^{(i)}_j$ almost surely. We define two events:
\begin{itemize}
    \item The event \( E \) indicates that in the first \( \frac{K}{4(1-\epsilon)} \) rounds, the erasure sequence results in at most \( \frac{K}{4} \) unerased rounds. Formally,
    \[
    E := \left\{ \sum_{t=1}^{\frac{K}{4(1-\epsilon)}} (1 - e_t) \leq \frac{K}{4} \right\},
    \]
    where \( e_t \) is the indicator variable for whether round \( t \) is erased, i.e.,
    \[
    e_t := \begin{cases} 
    1, & \text{if round } t \text{ is erased,} \\
    0, & \text{if round } t \text{ is not erased.}
    \end{cases}
    \]
    Since \( \epsilon > 0.5 \), the expected number of unerased rounds in \( \frac{K}{4(1-\epsilon)} \) trials is \( \frac{K}{4} \), ensuring that $\mathbb{P}[E] \geq \frac{1}{2}$.
    
    \item The event \( E'_i \) indicates that if the arm in the first round is erased, the agent does not select arm \( i \). Let \( \tilde{a}_0 \) be the arm chosen when the first round is erased. Formally,
    \[
    E'_i := \{\tilde{a}_0 \neq i\}.
    \]
    The probability of \( E'_i \) satisfies
    \[
    \mathbb{P}[E'_i] \geq \frac{1}{2} \text{ for at least } \frac{K}{2} \text{ arms.}
    \]
    This is derived from the condition
    \[
    \sum_{i=1}^{K} \mathbb{P}[\tilde{a}_0 = i] = 1 \implies \sum_{i=1}^{K} \left(1 - \mathbb{P}[E'_i]\right) = 1.
    \]
\end{itemize}

We will show that there is no policy $\pi$ that can make $\mathbb{E}[R_T(\pi, \nu_i)|E \cap E'_i]$ to be small for all $i$.

First, pick a set $I \subseteq [K]$ with $|I| = \frac{K}{2}$, and $\mathbb{P}[E'_i] \geq \frac{1}{2}$ for all $i \in I$.

Now, define the event $\mathcal{P}_i$:
\begin{itemize}
    \item \( \mathcal{P}_i \) is the event where arm \( i \) is not picked by the learner during any unerased round in the first \( \frac{K}{4(1-\epsilon)} \) rounds. Formally,
    \[
    \mathcal{P}_i := \left\{ \sum_{t=1}^{\frac{K}{4(1-\epsilon)}} (1 - e_t) \cdot \mathbf{1}\{a_t = i\} = 0 \right\},
    \]
    where \( \mathcal{H} \) is the history, \( a_t \) denotes the arm selected by the learner at round \( t \), and \( \mathbf{1} \) is the indicator function.
\end{itemize}

Let the set $A$ denote the indices of arms for which $\mathcal{P}_i$ holds, i.e., $A = \{i \in I | \mathcal{P}_i\}$, and let $B = I \backslash A$. We have that $A$ and $B$ are disjoint and their union is the set $I$. We consider the minimum worst-case probability of $i \in A$ conditioned on $E \cap E'_i$ under the distribution induced by instance $\nu_i$:
\begin{equation}
    \min_{\pi} \max_{i \in I} \mathbb{P}_{\nu_i}[i \in A | E \cap E'_i]. \label{eq:lb-min-max}
\end{equation}

We make the following observation (i):

After $\frac{K}{4(1-\epsilon)}$ rounds, there exist at least $\frac{|I|}{2} = \frac{K}{4}$ arms in $I$ for which $\mathcal{P}_i$ holds (hence, $|A| \geq \frac{|I|}{2}$ and $|B| \leq \frac{|I|}{2}$).

Consider two policies $\pi$ and $\pi^{nr}$ whose probability of selecting arms are the same deterministic function of the history, $\mathcal{H}$, with the following difference: $\pi$ receives the reward feedback and $\pi^{nr}$ assumes all generated rewards are $0$'s. It follows that, under the same erasure sequence and $E \cap E'_i$, for instance $\nu_i$, $\pi$ and $\pi^{nr}$ observe the same history and make the same selections until there is a reward feedback of $1$. Note that if there is a reward feedback of $1$, it means $i \not\in A$ which is a decision made by policies $\pi$ and $\pi^{nr}$  before the reward feedback of $1$. Therefore, the optimal value of Eq.~\ref{eq:lb-min-max} does not change if the learner does not observe rewards. Then, the learner can assume no reward feedback and proceed assuming that all previous rewards are zeros.

Additionally, since the reward feedback is eliminated, probability of $i\in A$ does not depend on the instance, reward outcomes and $\tilde{a}_0$; we can denote $\mathbb{P}_{\nu_i}[i\in A|E\cap E'_i]$ as $\mathbb{P}[i\in A | E]$. Assume there is a genie telling the learner the erasures of the first $K/(4(1-\epsilon))$ rounds in the beginning (note that this is a stronger erasure feedback; if the learner chooses to use only the erasure information of time slots $[0,t-1]$ at time $t$, it becomes equivalent to the original erasure feedback we consider). Consequently, the problem in Eq.~\ref{eq:lb-min-max} becomes equivalent to partitioning the set $I$ into the two sets $A$ and $B$ (possibly in a random way) ahead of time, with the goal of minimizing $\max_{i\in I}\mathbb{P}[i\in A | E]$.

The minimum value for $\max_{i\in I}\mathbb{P}[i\in A | E]$, and similarly, $\max_{i\in I}\mathbb{P}_{\nu_i}[i\in A|E\cap E'_i]$, is at least $1/2$ as $\sum_{i=1}^{|I|}\mathbb{P}[i\notin A | E]\leq |I|/2$ due to $|A|\geq |I|/2$ and $|B|\leq |I|/2$ (from observation (i)).

{Hence, for any deterministic policy $\pi$, there is an instance $\nu_{i^\star}, i^\star \in I$ such that conditioned on $E\cap E'_{i^\star}$, arm $i^\star$ is not picked for the unerased rounds by the learner in the first $K/(4(1-\epsilon))$ rounds with probability at least $1/2$.
For instance $\nu_{i^\star}$, the regret value increases by $1$ whenever arm $i^\star$ is not pulled, then, there is $i^\star \in I$ such that $\mathbb{P}\left[R_T(\pi,\nu_{i^\star})\geq \frac{K}{4(1-\epsilon)}|E\cap E'_{i^\star}\right]\geq 1/2$.} It follows that $\mathbb{E}[R_T(\pi,\nu_{i^\star})|E\cap E'_{i^\star}]\geq c\frac{K}{(1-\epsilon)}$. By non-negativity of regret, 
\begin{align}
    \mathbb{E}[R_T(\pi,\nu_{i^\star})] &\geq c\frac{K}{(1-\epsilon)}\mathbb{P}[E\cap E'_{i^\star}] \\ \nonumber &\stackrel{(a)}{=} c\frac{K}{(1-\epsilon)}\mathbb{P}[E]\mathbb{P}[E'_{i^\star}] \stackrel{(b)}{\geq} c/4 \frac{K}{(1-\epsilon)},
\end{align}
where $(a)$ follows from the fact that $E,E'_{i^\star}$ are independent, and $(b)$ follows since $\mathbb{P}[E] \geq \frac{1}{2}$ and $P(E'_{i^\star}) \geq \frac{1}{2}$ as $i^\star \in I$. 
We extend the lower bound to all policies, both deterministic and randomized, by applying Yao's Minimax Principle \cite{yao1977probabilistic}.
$\hfill{\blacksquare}$

\section{Proof of Theorem~\ref{thm:sos-sae}}
\label{apx:single-agent-sae-revised} 
\noindent{\textit{Proof.}
Define $E_i^{j}$ to be the event that for arm $j$ in batch $i$, the number of rounds until the successful transmission (i.e., the length of the transmit phase) is at most $\alpha = \lceil 2 \frac{\log{T}}{\log{(1/\epsilon)}} \rceil$, and define event $E$ to be the event that $E_i^{j}$ happens for all arms $j \in \{1,\cdots, K \}$ in all batches $i \in \{1,\cdots,\lfloor \log(T) \rfloor \}$. We have that 
\begin{align}
    \mathbb{P}[E] =  1-\mathbb{P}[E^C] \geq 1-\bigcup_{i,j} \mathbb{P}[{E_i^j}^C] \ge 1 - K\log (T) \epsilon^\alpha\geq 1-0.25/T.
\end{align}
Conditioned on $E$, the learner will wait at most $\alpha$ rounds to start collecting rewards for each arm in each batch. By concentration of sub-Gaussian random variables, we also have that the following event
$$E'=\{|\mu_a^{(i)}-\mu_a|\leq 2\sqrt{{\log (KT)}/{(2 \cdot m_i)}}\forall a\in A_i\forall i\in [\log T]\},$$
where $m_i$ is the number of pulls for each arm in the set of surviving arms, $A_i$, in batch $i$, occurs with probability at least $1-0.25/T$. Hence, we have that\vspace{-5pt}
$$\mathbb{P}[E\cap E']\geq (1-0.25/T)^2\geq 1-1/T.$$
In the remaining part of the proof we condition on the event $E \cap E'$. By the elimination criterion in Algorithm~\ref{alg:SoS}, and assuming $E \cap E'$, the best arm will not be eliminated. This is because the elimination criterion will not hold for the best arm as\vspace{-5pt}
\begin{align}
    \mu_a^{(i)}-\mu_{a^\star}^{(i)}&\leq \mu_a-\mu_{a^\star}+4\sqrt{\log(KT)/(2 \cdot m_i)}\nonumber \\
    &\leq 4\sqrt{\log(KT)/(2 \cdot m_i)} \forall a \forall i.
\end{align}
Now consider an arm $a$ with gap $\Delta_a>0$ and let $i$ be the smallest integer for which $4\sqrt{\log(KT)/(2 \cdot m_i)}< \frac{\Delta_a}{2}$. Then, we have that 
\begin{align}
    \mu_{a^\star}^{(i)}-\mu_a^{(i)}&\geq \mu_{a^\star}-\mu_a-4\sqrt{\log(KT)/(2 \cdot m_i)}>\Delta_a - \frac{\Delta_a}{2}\nonumber \\
    &>4\sqrt{\log(KT)/(2 \cdot m_i)}.
\end{align}
Hence, arm $a$ will be eliminated before the start of batch $i+1$. We also notice that from the definition of $i$, i.e., the smallest integer for which $4\sqrt{\log(KT)/(2 \cdot m_i)}< \frac{\Delta_a}{2}$, the value of $i$ can be bounded as
\begin{equation}\label{eq:bound_on_i_unknownerasure}
    i\leq \max\{1, \log_4(\frac{33\log(KT)}{\Delta_a^2})\}.
\end{equation}

By the exponential increase in the number of pulls of each arm, we get that until eliminated, conditioned on $E \cap E'$, arm $a$ will be pulled by the learner at most $$N_{T,a}\leq \sum_{j=1}^{i}(\alpha + 4^j) \overset{(1)}{\le} \sum_{j: \alpha > 4^j} 2 \alpha  + \sum_{j'=1}^{i'} 2 \alpha 4^{j'} \overset{(2)}{\le} c \alpha + 2 \alpha 4^{i'+1} \overset{(3)}{\le} c'(\alpha + \frac{\log(KT)}{\Delta_a^2}),$$
where 
\begin{itemize}
    \item[$(1)$:] follows from splitting the batches into two groups: those with $\alpha > 4^j$, for which $\alpha+4^j\leq 2\alpha$, and those with $\alpha\leq 4^j$ (reindexed as $j'$);
    \item[$(2)$:] follows by observing that the number of pulls in the first batch of the reindexed groups (i.e., $j'=1$) provides an upper bound on the first sum (i.e., $\sum_{j: \alpha > 4^j} 2  \alpha $);
    \item[$(3)$:] follows from applying the bound on $i$, see Eq.~\ref{eq:bound_on_i_unknownerasure} (rewritten as $i'\leq \max\{1,\, \log_4(33\log(KT)/(\alpha\Delta_a^2))\}$), 
\end{itemize} with $c, c'>0$ denoting some absolute constants. This results in a regret that is at most $c' (\alpha + \frac{\log(KT)}{\Delta_a})$. Summing the regret over all arms, we get that conditioned on $E \cap E'$ we have that
$$R_T\leq c'\left(\frac{K\log{T}}{\log(1/\epsilon)}+\sum_{a:\Delta_a>0}\frac{\log(KT)}{\Delta_a}\right).$$
This directly implies a worst-case regret bound of $O\left(\frac{K\log{T}}{\log{(1/\epsilon)}}+\sqrt{KT}\right)$. The proof is concluded by noticing that $\log(1/\epsilon)=O(1-\epsilon)$.

\end{document}